\def\year{2020}\relax
\documentclass[letterpaper]{article} % DO NOT CHANGE THIS
\usepackage{aaai20}  % DO NOT CHANGE THIS
\usepackage{times}  % DO NOT CHANGE THIS
\usepackage{helvet} % DO NOT CHANGE THIS
\usepackage{courier}  % DO NOT CHANGE THIS
\usepackage[hyphens]{url}  % DO NOT CHANGE THIS
\usepackage{graphicx} % DO NOT CHANGE THIS
\usepackage{xcolor} % REMOVE AFTERWARDS
\usepackage{amsmath}
\usepackage{graphicx}  % DO NOT CHANGE THIS
\title{Offline Learning for Planning: A Summary}
\author{Giorgio Angelotti,\textsuperscript{\rm 1,\rm 2} Nicolas Drougard,\textsuperscript{\rm 1,\rm 2} Caroline P. C. Chanel\textsuperscript{\rm 1,\rm 2}\\ % All authors must be in the same font size and format. Use \Large and \textbf to achieve this result when breaking a line
\textsuperscript{\rm 1}ANITI - Artificial and Natural Intelligence Toulouse Institute, Universit\'e de Toulouse,\\ %If you have multiple authors and multiple affiliations
41 Allées Jules Guesde, 31013 Toulouse - CEDEX 6, France\\
\textsuperscript{\rm 2}ISAE-SUPAERO, Universit\'e de Toulouse,\\
10 Avenue Edouard Belin, 31055 Toulouse - CEDEX 4, France\\
\{name.surname\}@isae-supaero.fr % email address must be in roman text type, not monospace or sans serif
}

\definecolor{persimmon}{rgb}{0.93, 0.35, 0.0}

\newcommand{\sachant}{\, \right| \left. \,}
\newcommand{\paren}[1]{\left( \left. #1 \right. \right)} 
\newcommand{\croch}[1]{\left[ \left. #1 \right. \right]}
\newcommand{\E}{\mathrm{E}}

\begin{document}

\maketitle

\begin{abstract}
The training of autonomous agents often requires expensive and unsafe trial-and-error interactions with the environment. Nowadays several data sets containing recorded experiences of intelligent agents performing various tasks, spanning from the control of unmanned vehicles to human-robot interaction and medical applications are accessible on the internet. With the intention of limiting the costs of the learning procedure it is convenient to exploit the information that is already available rather than collecting new data. Nevertheless, the incapability to augment the batch can lead the autonomous agents to develop far from optimal behaviours when the sampled experiences do not allow for a good estimate of the true distribution of the environment. Offline learning is the area of machine learning concerned with efficiently obtaining an optimal policy with a batch of previously collected experiences without further interaction with the environment. In this paper we adumbrate the ideas motivating the development of the state-of-the-art offline learning baselines. The listed methods consist in the introduction of epistemic uncertainty dependent constraints during the classical resolution of a Markov Decision Process, with and without function approximators, that aims to alleviate the bad effects of the distributional mismatch between the available samples and real world. We provide comments on the practical utility of the theoretical bounds that justify the application of these algorithms and suggest the utilization of Generative Adversarial Networks to estimate the distributional shift that affects all of the proposed model-free and model-based approaches.
\end{abstract}

\noindent Learning using a single batch of collected experiences is a statistical challenge of crucial importance for the development of intelligent agents, specially in scenarios where the interaction with the environment can be expensive, risky or unpractical. There are countless examples that fall in these categories: the training of unmanned aerial vehicles \cite{UAV}, self-driving cars \cite{selfdrivingcars}, medical applications \cite{medical}, Human-Robot interaction \cite{physio}. %Depending on the task, any mistake made by the agent can lead to catastrophic aftermaths. 
Several environments are so complex that a direct formulation of a model based on mere intuition is inappropriate and unsafe because, depending on the task, any mistake made by the agent can lead to catastrophic aftermaths. It is therefore necessary to infer the world dynamics from a batch of previously collected experiences. The said data set should be \emph{large} and \emph{diverse} enough for allowing useful information extraction. %as allow useful information to be easily extracted from it. 
%Unfortunately, sometimes the collection procedure notably requires an interaction with several human operators that usually are not available in standard laboratory conditions. Crowdsourcing using an online environment simulator has been recently exploited to overcome this hindrance \cite{crowdHRI,crowdHRI2,crowdHRI3}.

The process of learning an optimal policy can be mathematically formalized as the resolution of a Markov Decision Process (MDP) if the state of the system can be considered as fully observable and the action effects are non necessarily deterministic.
%or of a Partially Observable Markov Decision Process (POMDP) if the agent at each time step possesses only an incomplete knowledge of the system. 
This paper addresses the problems linked to the resolution of MDPs starting from a single batch of collected experiences by writing up a summary of the state-of-the-art methods on offline learning and planning, and outlining their pros and cons. When the data set is fixed the distributional shift between the true, unknown, underlying MDP and its best data-driven estimate can be non negligible and lead, on resolution, to bad performing policies. This discrepancy can be seen tightly linked to the uncertainty we possess about the model. Several offline learning baselines try to handle this issue or by constraining the policy or by reshaping the reward taking into account a local quantification of the said epistemic uncertainty and hence adapting the classical resolution paradigms \cite{tutorial}.\\ 
The document will be structured as follows:
\begin{enumerate}
    \item Firstly, a recap of MDP resolution with MDP planning algorithms, but also with reinforcement learning (RL) algorithms is proposed.
    \item Then, an intuitive description of offline model learning and batch RL is presented.
    \item And finally, a discussion is provided with comments on the theoretical guarantees for the performance of the listed baselines and suggestions for further improvements in this field, like resorting to Generative Adversarial Networks (GANs) to better estimate the underlying distributions.
\end{enumerate}

\section{A Review of MDPs}
\label{sec:mdp}
\noindent An MDP is formally defined as a tuple $M \stackrel{\mathrm{def}}{=} \left(S, A, T, r, \gamma, \mu_0\right)$ where $S$ is the set of states, $A$ the set of actions, $T: A\times S\times S\rightarrow \left[0, 1\right]$ is the state transition function defining the probability that dictates the evolution from $s \in S$ to $s' \in S$ after taking the action $a \in A$, $R: A\times S \rightarrow \left[R_{min}, R_{max}\right]$ with $R_{max}, R_{min} \in \mathrm{R}$ and $R_{max}>R_{min}$ is the reward function that indicates what the agent gains when it selects action $a \in A$ and the system state is $s\in S$, $\gamma \in \left[0, 1\right)$ is called the discount factor and $\mu_0: S \rightarrow \left[0, 1\right]$ is the initial probability distribution over states $s \in S$ at time $t=0$.
A policy is defined as a function that maps states to actions, such as $\pi: A\times S\rightarrow \left[0, 1\right]$; $\pi(a|s)$ can be interpreted as the probability of taking action $a \in A$ when being in the state $s \in S$. Time evolution is discrete and at every time step the agent observes the system, acts on the environment and earns a reward. The following definitions refer to an MDP with discrete $A$ and $S$ but they can be straightforwardly rearranged to address MDP with continuous states and actions spaces. 

Solving an MDP amounts to finding a policy $\pi^*$ which, $\forall s \in S$, maximizes the value function:
\noindent
\begin{equation*}
V_M^\pi(s)\stackrel{\mathrm{def}}{=}\mathrm{E}_{\substack{a_t \sim \pi(\cdot|s_t)\\ s_{t+1} \sim T(\cdot|s_t, a_t)}}\croch{\sum_{t=0}^{\infty}\gamma^t R(s_t, a_t) \sachant s_0 = s}.
\end{equation*}
The value function can also be expressed recursively as the fixed point of the Bellman operator:\\ 
%$V_M\left[\pi, s\right] = \sum_a \pi(a|s)\left( \sum_{s'}T(s'|a,s)\left(R(s,a)+\gamma V_M[\pi, s']\right)\right)$ 
\noindent
\begin{equation*}V_M^\pi(s) = \sum_{a\in A} \pi(a|s)\left(R(s,a)+ \gamma\sum_{s' \in S}T(s'|s,a)V_M^\pi(s')\right).
\end{equation*}
%\cite{sutton}.
We define also the Q-value function:\\
\noindent
\begin{equation*}
%Q^\pi(s, a)\stackrel{\mathrm{def}}{=} \mathrm{E}_{s'\sim T (\cdot| s,a)
%%s'\sim T\big(\cdot \big|s,\pi(a|s)\big)
%}\left[R(s,a) + \gamma V_M^\pi(s')\right]
Q^\pi(s, a) \stackrel{\mathrm{def}}{=} R(s,a) + \gamma \sum_{s' \in S}  T \paren{ s' \sachant s,a} V_M^\pi(s')
\end{equation*}
and we notice that $V_M^\pi(s) = \mathrm{E}_{a \sim \pi(\cdot|s)} \croch{Q^\pi(s, a)}$.

\subsection{Resolution Schemes}
\noindent With basic planning algorithms like Value or Policy Iteration where the contraction property of the Bellman operator is exploited, one can compute a value $V$ and a policy $\pi$ that iteratively converge to $V^*$ and $\pi^*$, respectively \cite{sutton,kolobov}. These algorithms require to store in memory the whole state space.
However, the application of the Bellman operator demands that all the functions that compose the MDP are known. What can we do, for instance, if the transition function is unknown? %and reward

\subsubsection{Model-free Approaches}
In such scenarios, temporal difference (TD) schemes like \textit{Q-learning} \cite{QL} can be applied. In Q-learning the Q-function is computed iteratively by minimizing the TD error using sampled transitions $(s, a, r, s')$.
%\noindent
%\begin{align}Q_{t+1}(a,s)\leftarrow (1-\alpha_t) Q_t(a,s) + \alpha_t \left(R(a,s') + \max_{a'} Q_t(a', s')\right)
%\end{align}
Q-learning is a model free RL algorithm because 1) it does not require an a-priori knowledge of the model, 2) it exploits a growing batch of sampled experiences. %When an appropriate learning rate is applied and the $(s,a)$ transitions are sufficiently visited then Q-learning is proved to converge to the optimal Q-value with probability $1$ \cite{sutton}.
Another popular model free approach is based on Policy Gradients \cite{REINFORCE} which maximizes an estimate of the value function with respect to the policy where the expected value over the transition distribution is replaced by sampled transitions.

Policy Gradients methods serve as the base for the Actor-Critic architecture by which the variance of the gradient with respect to the policy is reduced either by replacing the cumulative reward with an estimate of the Q-value or by subtracting from it an estimate of the value function \cite{AC}. The module that compute Q-value and value function estimates is called the Critic and the one that computes $\pi$ thanks to the Policy Gradient method is named the Actor.

\subsubsection{Model-based Approaches}

\noindent Another route to follow is that of first using a batch of previously sampled experiences to obtain both $\hat{T}$ and $\hat{R}$ which are respectively estimates of the transition function $T$ and of the reward function $R$ of the unknown MDP. And then, to directly execute a planning algorithm or to use $\hat{T}$ as a generative model of new fictitious experiences $(s, a, r, s')$ and subsequently apply a model-free technique using the new data set augmented with the artificial transitions. Such a scheme was first mentioned in the Dyna-Q algorithm \cite{sutton}, even though it was prescribed to be used in combination with periodical further explorations of the environment. 
{Regarding the MDP planning literature, techniques which exploit heuristic guided trial-based solving have been created to address large finite state spaces. Amongst all \textit{Upper Confidence bounds applied to Trees} (UCT) and more recently, PROST \cite{UCT,PROST}. Such algorithms could be applied to the estimated MDP generative model, using $\hat{T}$ and $\hat{R}$.}

Notice that when new data is generated the optimal policy of the MDP defined by $\hat{T}$ can be different from the one of the original MDP since they really define two different decision processes. The authors of the work \cite{model} questioned the advantage of using a model to generate fictitious data over working directly on the batch with model free algorithms. Model-based techniques are normally more data efficient than model free competitors since probably the model learning stage can capture more easily the characteristics needed to estimate the Q-value and value function. However, in that paper it is empirically displayed that an appropriately fine-tuned model-free algorithm can achieve a superior data efficiency performance.

%\begin{comment}
%\subsection{Relational Planning}
%Planning is classically applied on MDPs with states indicated by categorical variables which represent a compact cluster of features: e.g. one can express the position of a pen as ``in the box" or by its three dimensional coordinates. On these terms the model dynamics is given as a set of probabilistic transition \emph{relational rules} which are imposed by a model engineer or learnt statistically from a batch using algorithms that are often based on Relational Decision Trees. An exhaustive review of Relational Planning and Action Model Learning has been summarized by \cite{relplan}. In this paper we will not focus on relational planning but we will rather shed light on states which are designated by sets of features that can be both continuous, both categorical. This because nowadays robots can get high dimensional observations from their sensors and a translation to relational variables is not always straightforward.
%
%Dimensionality reduction, state aggregation and state representation learning are extremely important since the computational time of a run of Value/Policy iteration is polynomial in $|S|$. Moreover, function approximators usually benefit from an information preserving reduction of the dimensionality of the inputs. A good review of the subject has been written up by \cite{Natalia}.
%\end{comment}
\subsection{Function Approximators}
\noindent When the state or the action space has the cardinality of the continuum a tabular representation of policies and value functions is unfeasible. If the states are characterized by continuous feature vectors, planning algorithms are not applicable without a preliminary discretization. 
% see Sample-Based Planning for Continuous Action Markov Decision Processes. Chris Mansley, Ari Weinstein, Michael Littman
% In this paper, we present a new algorithm that integrates recent advances in solving continuous bandit problems with sample-based rollout methods for planning in Markov Decision Processes (MDPs). Our algorithm, Hierarchical Optimistic Optimization applied to Trees (HOOT) addresses planning in continuous-action MDPs. Empirical results are given that show that the performance of our algorithm meets or exceeds that of a similar discrete action planner by eliminating the problem of manual discretization of the action space.
%%%%%%%%
\cite{munosmoore} proposes a variable resolution discretization of $S$ assuming that a perfect generative model is available. The latter enables to split recursively the feature space where more control is required preventing an unforgivable loss of resolution in the transition function of the aggregate MDP. Even though the variable resolution scheme provides a more efficient splitting criterion than an uniform grid, it does not manage to escape from the curse of dimensionality. Conversely, some promising attempts have been fulfilled in the case of a continuous action space and finite state space \cite{littman}. Resorting to the Universal Approximation Theorem \cite{balasz} it has been found practical to use function approximators in order to estimate the policy and the value functions. The increase in computational power of the last decade gave birth to a rich community of scientists and engineers who use function approximators with thousands of parameters such as neural networks. Model-free algorithms using neural networks are Deep Q-learning Networks \cite{DQN}, Policy Gradients \cite{REINFORCE,AC} and their subsequent improvements \cite{Rainbow} \cite{TRPO,A3C,D4PG,SAC} that led to development of agents which achieved better than human performances in games like Go \cite{alphago}, Chess \cite{chess}, and also some video games of the ATARI suite.

In model-based settings, approximators usually need the specification of a prior distribution for $T$ which is often chosen Gaussian since these algorithms are usually applied to problems driven by a deterministic dynamics or to problems whose intrinsic stochasticity can be thought being induced by a Gaussian distribution in some latent space \cite{pilco,chua,planet,simple,dreamer}. The latter is a strong limitation of these approaches since, more often than not, taking decisions under uncertainty amounts to deal with multi-modal transition distributions that would be poorly described by a normal distribution.

\section{Single Batch Learning}
\label{sec:batch}
As we have stated in the introduction, learning from a single batch of collected experiences is a necessity of compelling importance for a safe, cost limited and data efficient development of intelligent agents. We will see that several algorithms which constrain the optimal policy obtained with RL or planning tools to one that does not drive the agent to regions of $S \times A$ that have been poorly sampled in the data set lead to more effective policies than the one used to collect the batch. Usually the results are also better than the one obtained with a policy that has been calculated by straightforwardly applying the schemes listed in Section \ref{sec:mdp}.

The utilization of function approximators to estimate the value functions using a single batch requires theoretical delicacy since many convergence guarantees do not stand. In the paper \cite{Chen} the authors realized that usually two fundamental assumptions are implicitly required in order for the following algorithms to work:
\begin{enumerate}
\item mild shifts between the distributions of the real world and the one inferred from the data in the batch,
% A class of functions allowing to make good approximations
\item conditions on the class of candidate value-functions stronger than just the membership of the optimal Q-value to this function class.
\end{enumerate}
Related to those points, \cite{Chen} explores the notion of \emph{concentratability} coefficient \cite{munos}, which is hereafter recalled. %Given an MDP M, if there exist a policy $\pi$ and a $h\geq0$ such as we can define an \emph{admissible distribution} in this way: $\nu(s,a)\stackrel{\mathrm{def}}{=} P \left( s_h= s, a_h = a | s_0 \sim \mu_0, \pi\right)$, given also a \emph{data distribution} $\mu_{\mathcal{B}}(s,a)$ which is \ga{approximation of the data distribution under the true dynamics which however assumes the samples to be i.i.d. (hence uncorrelated)},they define a \emph{concentratability} coefficient \cite{munos} $C < \infty$ that for any admissible $\nu$

$\forall (s,a) \in S \times A$ and $\forall h \geq0, \forall\pi$:
\begin{equation*}
\frac{P \left( s_h= s, a_h = a | s_0 \sim \mu_0, \pi\right)}{\mu_{\mathcal{B}}(s,a)}\leq C,
\end{equation*}
where $\mu_{\mathcal{B}}$ is the probability distribution that generated the batch assuming that the transitions are independent and identically distributed. The existence of $C$ ensures that any attainable distribution of state-action pairs is not too far away from $\mu_{\mathcal{B}}$.
The main result reported in \cite{Chen} is that not constraining $C$ precludes sampled efficient learning even with ``the most favourable" data distribution $\mu_{\mathcal{B}}$.

Rather than focusing on the practical implementation of the different methodologies, which as we will see is often approximate due to the intractability of the terms present in the derived theoretical bounds, we aim to perform a simple yet comprehensible adumbration of the ideas that support their development. With this in mind we are going to neglect implementation related technicalities and sketch the theoretical foundations of single batch learning algorithms.

\subsection{Constraints for Model-free Algorithms}
The first successful applications of offline learning for planning and control with function approximators are very recent \cite{BCQ,BCQ2}. In these works the authors showed that performing Q-learning to solve a finite state MDP using a fixed batch $\mathcal{B}$ leads to the optimal policy $\pi_{\mathcal{B}}^{*}$ for the MDP $M_{\mathcal{B}}$ whose transition function is the most likely one with respect to the transitions $(s, a, r, s') \in \mathcal{B}$.
%Extrapolation error can be attributed to a
%mismatch in the distribution of data induced by the policy
%and the distribution of data contained in the batch.
More often than not, the optimal policy for $M_{\mathcal{B}}$ performs poorly in the true environment.
The discrepancy between the transition function of the original process and the one learnt from the batch will be the key element of the following discussion.

Indeed, the \emph{extrapolation error} for a given policy $\pi$: \noindent
\begin{equation*}
\epsilon(s,a)\stackrel{\mathrm{def}}{=}Q^{\pi}(s,a)-Q_{\mathcal{B}}^{\pi}(s,a)
\end{equation*} 
defined as the difference between the Q-value function of the real MDP and the Q-value function of the most likely MDP learnt from the batch could be computed with an operator similar to the Bellman's one:

%\noindent
%\begin{eqnarray}
%    \epsilon(s,a)&=&\sum_{s' \in S} \big( T(s'|s,a)-T_{\mathcal{B}}(s'|s,a) \big) \Big(R(s, a)+\nonumber\\
%    && + \gamma \sum_{a' \in A}\pi(a'|s')Q_{\mathcal{B}}^{\pi}(s', a')\Big)\nonumber\\
%    && + T(s'|s,a)\gamma\sum_{a' \in A} \pi(a'|s')\epsilon(s',a')
%\end{eqnarray}

\begin{eqnarray*}
    \epsilon(s,a)&=& \gamma \sum_{s' \in S} \Big[\big( T(s'|s,a)-T_{\mathcal{B}}(s'|s,a) \big) V_{\mathcal{B}}^{\pi}(s') \nonumber\\
    && + T(s'|s,a) \sum_{a' \in A} \pi(a'|s')\epsilon(s',a')\Big]
\end{eqnarray*}
%Am I wrong? 
% \ga{Ah I see, I had copied this equation from the paper}

%\noindent
%\begin{align}
%    \epsilon(s,a)=\sum_{s'}\left(T(s'|a,s)-T_{\mathcal{B}}(s'|a,s)\right)\Big(r(s, a, s')+\nonumber\\+\gamma \sum_{a'}\pi(a'|s')Q_{\mathcal{B}}^{\pi}(s', a')\Big)+T(s'|a,s)\gamma\sum_{a'}\pi(a'|s')\epsilon(s',a')
%\end{align}

The authors noticed that the extrapolation error is a function of divergence between the true transition distribution and the one estimated from the batch along with the error at succeeding states. Their idea is then to minimize the error by constraining the policy to visit regions of $S\times A$ where the transition distributions are similar.
Henceforth, they modified Q-Learning and Deep Q-Learning algorithms to force the new ``optimal" policy to be \emph{not so distant} from the one that was used during the collection of the batch. They train a generator network that gets as an input a state $s$ to estimate the batch generating policy and then allow for a small perturbation around it. The magnitude of the perturbation is an hyperparameter. In this way, they obtain a policy that always achieves better performance than the one used during the batch collection. This algorithm is called Batch Constrained Q-Learning (BCQ).

In a subsequent work, it has been shown that the error in the estimation of the Q-value with neural networks is generated by the back-up of poor estimates of the Q-value that comes from regions $S \times A$ that were badly sampled in $\mathcal{B}$ \cite{BEAR}. To contrast the accumulation of the error, the authors developed the Bootstrapping Error Accumulation Reduction (BEAR) algorithm which, exploiting the notion of distribution concentratability, manages to constrain the improved policy to the support of the one that generated the batch. Strictly speaking, they blame the back-up of Q-value estimates of states with Out Of Distribution (OOD) actions for increasing the extrapolation error. They should blame for OOD state-action transitions, but in offline Q-learning the Q function is computed only at states that are in the replay buffer. This constraint is softer than the one imposed by BCQ and it has been showed to provide better results.

When BEAR and BCQ are applied on batches generated with a random policy, they can eventually perform worse than Deep Q-learning naively applied using the batch as a fixed replay buffer. In these cases, if the data set is big enough, there are not many OOD actions \cite{BEAR}. Probably, enforcing a constraint as done in BEAR and BCQ will provide a too little window for policy improvement.

In the same year, yet another inspiring paper about offline reinforcement learning was published \cite{BRAC}. The authors of the latter showed that any policy constraining approach like BCQ, BEAR and KL-Control \cite{KLC} can be obtained as a special case of their Behaviour Regularized Actor Critic (BRAC) algorithm.

The general idea is to either 1) penalize the value function estimated by the actor or 2) regularize the policy generated by the critic by a distance in probability space between the batch collector policy $\pi_\mathcal{B}$ and the currently evaluated one, as:
\noindent
\begin{align*}
V_D^{\pi}(s) =  \sum_{t=0}^{\infty}\gamma^t \mathrm{E}_{\substack{a_t \sim \pi(\cdot|s_{t})\\s_{t+1} \sim T(\cdot|s_t,a_t)}}\Big[r(s_t,a_t)\nonumber\\-\alpha D\big(\pi(\cdot|s_{t}),\pi_{\mathcal{B}}(\cdot|s_{t})\big)|s = s_0\Big]
\end{align*}
where $D$ is a distance function in probability space (e.g. Kernel MMD, Kullback-Leibler, Total Variation, Wasserstein, etc) and $\alpha$ is an hyperparameter.
While the policy regularized learning objective of the actor maximizes the following criterion:

\begin{equation*}
\displaystyle \underset{(s,a,r,s') \sim \mathcal{B}}\E \croch{
\underset{a\sim \pi(\cdot|s)}\E\croch{Q(s,a)}
-\alpha D\big( \pi(\cdot|s),\pi_{\mathcal{B}}(\cdot|s) \big)}
\end{equation*}
Their results showed that overall value penalization works better than policy regularization and the distance  $D$ that provides the best performing policy is the Kullback-Leibler divergence.

%In their practical implementation both BCQ, BEAR (and BRAC) use an average Q-value (the minimum) \cc{average or minimum??} \ga{Brac the minimum} over an ensemble of Q-value networks to reduce the prediction error.
In their practical implementation both BCQ and BEAR use an average Q-value over an ensemble of Q-value networks to reduce the prediction error. In BRAC, the minimum Q-value over an ensemble of Q-value networks is used.

\subsection{Extrapolation Error Reduction with Random Ensembles of Q-value Networks}
\noindent In parallel to the previous studies, the authors of \cite{rem} empirically demonstrated that the stabilization of Deep Q-learning networks using a single data set can be achieved by training at the same time a multitude of different Deep Q-value Networks with their weights differently initialized. During training the final estimate of the Q-value will be a normalized \emph{random} linear combination of the output of the intermediate Q-functions, while in the end they will just consider as the final Q-value estimate their average. The linear combination step is equivalent to a Dropout layer in a neural network. By doing so the final output will be stabler and if a network will be more affected than another by an OOD action back-up, the final average over the random ensemble will likely mitigate this error. The authors called this neural network architecture Random Ensemble Mixture (REM).

\subsection{Generative model learning for Model-based approaches}
\noindent Steps forward in the development of model based RL using a single batch have been done respectively in MOPO and MOReL \cite{MOPO,MOReL}. In both cases, a generative model is first learnt from the batch and then used to create new transitions. On the augmented data set then a model-free algorithm is applied. In this fashion, since we can use the generative model to ``explore" the $S \times A$ space, the error in the Bellman back-up will not be induced directly by ill sampled regions but by the \emph{epistemic error} of the model.

Intuitively, the more chaotic is the underlying system, the greater will a trajectory generated by the learnt model diverge from a real one given the very same starting state distribution and an identical sequence of actions to apply. Broadly speaking, as described below, the two methods build a penalized (MOPO) and pessimistic (MOReL) MDP whose optimal policies are encouraged to visit regions of $S\times A$ where the epistemic error is expected to be little (MOPO), or areas that would be likely to be sampled by the same distribution dynamics that generated the batch (MOReL).

\subsubsection{Model Error Penalized MDP}
In MOPO, defining as $\eta_{M}[\pi]\stackrel{\mathrm{def}}{=}\mathrm{E}_{s\sim\mu_0} \croch{V^{\pi}_{M}(s)}$ as the performance of a policy for the MDP $M$, a theoretical bound for $\eta_{M}[\pi]-\eta_{\hat{M}}[\pi]$ is recovered.
In particular, they show that:
\noindent
\begin{align}\label{eq:mopo}
\eta_{M}[\pi]\geq \mathrm{E}_{(s,a) \sim \rho_{\hat{M}}^{\pi}}\Big[r(s,a) - \frac{\gamma}{1-\gamma}\max_{s'}\big(V_{M}^{\pi}(s')\big)\nonumber\\  \cdot D_{TV}\left(T(\cdot|s,a),\hat{T}(\cdot|s,a)\right)\Big] = \eta_{\tilde{M}}\left[\pi\right]
\end{align}
where $\rho_{\hat{M}}^{\pi}$ is the discounted state-action distribution of transitions along the Markov Chain induced by $\hat{T}$ and $\pi$ starting from the initial state distribution $\mu_{0}$.
%\ga{There is no $-$ sign, the $D_{TV}$ multiplies the last term of the previous row. Everything in the right hand side is the definition of performance of the reward penalized MDP $\eta_{\tilde{M}}$ because you have the original reward $r$ minus the penalty term and the average is over $\rho_{\hat{T}}^{\pi}$}

The right hand side is the performance of the MDP $\tilde{M}$ whose dynamics is driven by $\hat{T}$, but with reward function penalized by a term which is directly proportional to the total variation distance between the true and the inferred transition functions. Since both $D_{TV}$ and $\max_{s'}V^{\pi}_M(s')$ are unknown, in the practical implementation the penalty is replaced by $\lambda u(s,a)$ where $\lambda$ is an hyperparameter and $u: S \times A \rightarrow \left[0, +\infty\right)$ such that $u(s,a)\geq D_{TV}(T(\cdot|s,a),\hat{T}(\cdot|s,a))$ $\forall (s,a) \in S \times A$.
Therefore finding the optimal policy for the penalized MDP amounts to obtaining the policy that maximizes the lower bound on $\eta_{M}$.

Notwithstanding, we believe that their bound is greatly dependent on the choice of a proper hyperparameter $\lambda$ and function $u$, which is not trivial for stochastic MDPs while it can be appropriately approximated by the covariance of a Gaussian Process for deterministic environments like the one used as test-cases in their paper. Moreover, if the penalization is too big the lower threshold will be likely of little use. Imagine the extreme situation where $\forall (s,a) \in S\times A$, $r > 0$ and $r - \lambda u < 0$. In this case $\eta_M[\pi]>0$ $\forall \pi$ trivially, while, calling $\tilde{M}$ the reward penalized MDP with dynamics driven by $\hat{T}$, $\eta_{\tilde{M}}[\pi]<0$. Therefore, maximizing the bound will not necessarily lead to a policy that works better than chance on the real MDP.

In \cite{MOPO} the performance of a policy is defined as the expected value of $V^{\pi}_{M}$ over the starting state distribution $\mu_0$. This starting state distribution is then interpreted as the distribution of states in the batch. However, the latter could be much different from the true starting state distribution if the batch is of modest size. 

Therefore a more robust definition could be $\eta_{M}[\mu_{M}^{\pi},\pi]\stackrel{\mathrm{def}}{=}\mathrm{E}_{s \sim \mu_{M}^{\pi}}\left[V^{\pi}_{M}(s)\right]$, where $\mu_{M}^{\pi}$ is the stationary distribution of states (if it exists) for the MDP $M$ with dynamics dictated by the policy $\pi$.
Using this new definition, the lower bound on $\eta_{M}[\mu_{M},\pi]$ acquires an extra term dependent on the difference $\Delta_{\hat{M},M}^{\pi}(s)=\mu_{\hat{M}}^{\pi}(s)-\mu_{M}^{\pi}(s)$.
Indeed, the performance of a policy in the true MDP can be expressed as:
\noindent
\begin{eqnarray*}
\eta_{M}\left[\mu_{M}^{\pi},\pi\right] & = &\mathrm{E}_{s \sim \mu_{M}^{\pi}}\left[V_{M}^\pi(s)\right] \nonumber \\ 
& = & \mathrm{E}_{\mu_{M}^{\pi}}\croch{V_{M}^\pi(s)}-\int_{S}d\mu(s)\Delta_{\hat{M},M}^{\pi}(s)V_{M}^\pi(s) \nonumber
\end{eqnarray*}
%\underset{s \sim \mu_{\hat{M}}^{\pi}}
where, $d\mu(s)$ is a measure over the state space. It is then possible to obtain the same bound of Equation (\ref{eq:mopo}) but with the extra term dependent on the integral over the state space of the discrepancy between the stationary distributions:
\noindent
\begin{eqnarray*}
    \eta_{\hat{M}}\left[\mu_{\hat{M}}^{\pi},\pi\right] - \eta_{M}\left[\mu_{M}^{\pi},\pi\right] = \nonumber \\=  \mathrm{E}_{s \sim \mu_{\hat{M}}^{\pi}}\left[V_{\hat{M}}^\pi(s)\right] - \mathrm{E}_{s \sim \mu_{M}^{\pi}}\left[V_{M}^\pi(s)\right] = \nonumber \\
    = \mathrm{E}_{s \sim \mu_{\hat{M}}^{\pi}}\left[V_{\hat{M}}^\pi(s) - V_{M}^\pi(s)\right]+\int_{S}d\mu(s)\Delta_{\hat{M},M}^{\pi}(s)V_{M}^\pi(s) \nonumber
\end{eqnarray*}
\noindent where the expected value over the distribution $\mu_{\hat{M}}^{\pi}$ is similar to the one computed in \cite{MOPO} but with $\mu_0 = \mu_{\hat{M}}^{\pi}$. Therefore the right hand side can be bounded from below:
\begin{eqnarray*}
    \eta_{\hat{M}}\left[\mu_{\hat{M}}^{\pi},\pi\right] - \eta_{M}\left[\mu_{M}^{\pi},\pi\right] \leq \frac{\gamma}{1-\gamma}\max_{s'}\left(V_{M}^{\pi}(s')\right) \nonumber \\ \cdot\mathrm{E}_{(s,a)\sim \rho_{\hat{M}}^{\pi}}\left[D_{TV}\left(T(\cdot|s,a),\hat{T}(\cdot|s,a)\right)\right] \nonumber \\ +\int_{S}d\mu(s)\Delta_{\hat{M},M}^{\pi}(s)V_{M}^\pi(s)
\end{eqnarray*}
where $\rho_{\hat{M}}^{\pi}$ now is the discounted state-action distribution of transitions along the Markov Chain induced by $\hat{T}$ and $\pi$ starting from the stationary distribution $\mu_{\hat{M}}^{\pi}$.
\\\noindent Exploiting the definition of penalized MDP $\tilde{M}$:
\begin{eqnarray*}
 \eta_{M}\left[\mu_{M}^{\pi},\pi\right] \geq \eta_{\tilde{M}}\left[\mu_{\hat{M}}^{\pi},\pi\right] -\int_{S}d\mu(s)\Delta_{\hat{M},M}^{\pi}(s)V_{M}^\pi(s)
\end{eqnarray*}
The latter can again be bounded from above by plugging in the absolute value of $\Delta$ and the max of $V$ over the state space:
\begin{eqnarray*}
    \eta_{M}[\mu_{M}^{\pi},\pi] & \geq & \eta_{\tilde{M}}[\mu_{\hat{M}}^{\pi},\pi]\\
    &&-\max_{s'}V_{M}^{\pi}(s')\int_S d\mu(s)\lvert\Delta_{\hat{M},{M}}^{\pi}(s)\rvert \nonumber
\end{eqnarray*}

It is remarkable that now the optimal policy for $\tilde{M}$ does not necessarily maximizes the bound. Assuming that our algorithm is monotonically improving the policy, it could be then convenient to stop it earlier and settle for a sub-optimal policy which in turn maximizes the bound. It's all about balancing the trade-off between the optimality condition for $\tilde{M}$ and the discrepancy within the stationary distributions. The newly added term is unfortunately intractable due to the lack of knowledge about the MDP.

In the implementation of MOPO new trajectories are generated starting from states already present in the batch up to $h$ following time steps. Ablation experiments have shown that the roll-out horizon $h$ is indeed required to obtain good results. We suspect that the state distributional shift that was neglected is to be blamed for the occurrence of the behaviour. Generating data that are not so far away from ones in the batch prevents the accumulation of model error, but this theoretical aspect, even if already mentioned in \cite{MBPO}, should not affect the bound that aims to be valid on any uncertainty penalized MDP independently of other factors.

\subsubsection{Pessimistic MDP}
The authors of MOReL define an MDP with an extra absorbing state $y$. The state space of the pessimistic MDP is $\tilde{S} = S \cup \{y\}$, while the transition function
\noindent
\begin{align*}
\tilde{T}(s'|s,a) =
\begin{cases}
\delta_{s',y} &\text{ if 
$D_{TV}\left(\hat{T}(\cdot|s,a),T(\cdot|s,a)\right)>\theta$,}\\
\delta_{s',y} &\text{ else if $s = y$,}\\
\hat{T}(s'|s,a) &\text{ otherwise.}
\end{cases}
\end{align*}
The reward function $R$ is identical to the original one except for $y$: $R\left(y, a\right) = -\kappa$     $\forall a \in A$.\\
$\theta$ is a freely chosen threshold and $\kappa>>0$ is a penalty.
Essentially if the model error is greater than $\theta$ the agent will end up for sure in the strongly penalized absorbing state. Therefore any optimal policy for the pessimistic MDP will try to avoid transitions for which the model error is high.
The optimal policy $\hat{\pi}$ when applied on the real MDP bounds from above the performance of the optimal policy of the true MDP.
\noindent
\begin{align*}
\small
    %\frac{4 R_{max}}{1-\gamma} \left(D_{TV}\left(\mu_0,\hat{\mu}_0\right)+\frac{\gamma}{1-\gamma}\max_{s',s,a}D_{TV}(\hat{T},T)+\mathrm{E}\left[\gamma^{T_{\mathcal{U}}^{\pi*}}\right]\right) \nonumber\\
    \frac{4 R_{max}}{1-\gamma} \left(\zeta(\mu_0,\hat{\mu}_0,T,\hat{T})+\mathrm{E}\left[\gamma^{T_{\mathcal{U}}^{\pi*}}\right]\right) \nonumber\\
    \geq \eta_{M}[\mu_0, \pi^*]-\eta_{M}[\mu_{0},\hat{\pi}]
\end{align*}
\noindent with,
\begin{equation*}
    \zeta(\mu_0,\hat{\mu}_0,T,\hat{T}) = D_{TV}\left(\mu_0,\hat{\mu}_0\right)+\frac{\gamma}{1-\gamma}\max_{s,a}D_{TV}(\hat{T},T) \nonumber
\end{equation*}

The two performances are similar if the $D_{TV}$ between the real starting state distribution and the one inferred from the batch is negligible, if the maximum model error is little, and also, if the expected value of the first hitting time of the absorbing state while applying the policy $\pi^*$ in the pessimistic model $\gamma^{T_{\mathcal{U}}^{\pi*}}$ is small.

Again, in a practical implementation the choice of good estimators of the epistemic error, of the distributional distance between starting states, and also, of the first hitting time is of crucial importance.
In the large batch regime the authors neglect the first two terms and focus only on the expectation of the first hitting time which can be bounded from the above by the \emph{discounted} distribution of visits to $(s,a) \in \mathcal{U}$, where $\mathcal{U}$ represents the \emph{unknown} state-action pairs that lead to the absorbing state, when applying $a \sim \pi^*$.%, namely the \emph{unknown} state-action transitions that lead to the absorbing state. 
The late distribution can be in turn bounded by a term proportional to the support mismatch of the distribution of states that were never sampled in the original data set.

\section{Discussion}
\label{sec:discussion}

\subsection{Theoretical bounds and function approximators}
\noindent The theoretical bounds which justify the creation of the previously listed algorithms rely either on a penalty or on a regularization term proportional to a sort of uncertainty that obnubilates our knowledge about the underlying system.
Sometimes the penalty is expressed as an estimate of the epistemic model error, other times as a difference between starting or stationary state distributions, finally it can be quantified as Out Of Distributions (OOD) state-action pairs with respect to the policy used during the collection of the data set.

The penalty or regularization term is often proportional to an upper bound of the value function or to a free hyperparameter.
As we have seen the latter statement implies that when this constant is too big the intractable performance of a policy on the real MDP is bounded by a tractable term which unfortunately will be of little use.

Only REM stabilizes the accumulation of the error in Q-learning thanks to a constraintless weighted random ensemble average. Despite its nicety, the stabilization is not properly a goal-oriented correction to the deviation of the optimal Q-value estimated using a single batch from the real one.

Model-based approaches also learn a function $\hat{R}$, however there is no term linked to the uncertainty in the evaluation of the reward in the batch penalized resolution scheme for offline learning algorithms. We believe that such a term proportional to the reward error is not truly necessary since we expect it to be stemmed from the same regions of $S \times A$ that are badly sampled in the data set $\mathcal{B}$ and considering that the penalization is already applied on the reward or value functions.

Finding a proper estimator of the \emph{errors} is not trivial. The algorithms were often tested on deterministic environments where a reasonable estimator of the model error can be achieved by the maximal variance in between an ensemble of different Gaussian models.
Since it's reasonable to expect that the model error will be high in regions that were ill-sampled in the batch, another way to measure it could be getting an estimate of the probability that a given $(s,a)$ could have been generated by the same process that gave birth to the batch. Therefore estimating the probability distribution of $(s,a,s')$ in the true MDP with policy $\pi_{\mathcal{B}}$ is a priority.

The most practical way of learning a probability distribution function without a prior could be to use a GAN \cite{GAN}.
A GAN is comprised of a Generator and a Discriminator. The first is a neural network which receives random noise as an input and generates an output with the same shape of the data in a training set. The second gets an input with the correct shape and provides as output a real number.
While training the Discriminator tries to identify which data was present in the training batch between samples that really populate it and the output of the generator. The higher the output of the Discriminator on a sample will be, the most likely that sample will be in the batch if the Discriminator is well trained.
At the same time the goal of the Generator will be that of fooling the Discriminator.
The loss functions minimized during the training are peculiar of a min-max game.
Sophisticated GANs architecture can use a well trained Generator to build fake samples that could fool even a human. A striking example is StyleGAN2 by NVIDIA \cite{StyleGAN2} which can generate high quality dimensional pictures of people that do not really exist.

We believe that the use of a GAN's Discriminator trained on $\mathcal{B}$ to obtain an estimate of the log-likelihood of a transition $(s,a,r,s')$ with respect to the unknown transition distribution should be a promising venue for penalizing the reward and/or the value functions with a more pertinent estimator of the distance between the true distribution of data and the one we can infer from a single batch. In this way we may be able to recover an informative quantity about the distributional shifts in a non parametric way that is independent of any possible prior and might, in principle, also work for systems driven by a stochastic time evolution. Doing so we would drop off the Gaussian assumption that has been so far used in almost all of the model-based techniques.

However, GANs have some weak spots: the training is unstable because the loss function is not convex, the procedure takes time, and they suffer from mode collapse. The latter is maybe the most problematic issue since when the unknown latent distributions is multi-modal the Generator may focus on building up samples that benefits from characteristics that are typical only of a little slice of the whole set.
Since the Discriminator is trained alongside the Generator, it will learn to recognize samples that are typical of that specific mode.
Several approaches to mitigate mode collapse \cite{madgan} and training instability \cite{wgan} have been attempted so far, but the issues can be still considered unsolved.

\subsection{Off-policy Evaluation}
It is necessary to find statistically robust methods which are able of estimating how well an algorithm will run in the real world without interacting with it. Off-policy evaluation is an active field which would require a summary of its own. Recent approaches utilize optimized versions of Importance Sampling to estimate the unknown ratio between stationary distributions of states under dynamics driven by different policies. Recent works propose to create a sort of Discriminator and optimize a min-max loss function to serve this purpose \cite{horizon,gendice}.

The application of a min-max optimization loss function in the field strengthens our intuition that the implementation of a GAN anyhow in the estimation of the distributional shift might be useful.

\subsection{Batch Quality and Size Scalability}
 It is of compelling necessity to develop and test the baselines in common environments using the same batches to shine a light onto the change in performance of the different paradigms when the quality (and the variety) and the size of the available data increase. D4RL (Datasets for Deep Data-Driven Reinforcement Learning) is a collection of data sets recorded using policies of different qualities (random, medium, expert) on the typical benchmarking environments used by the RL community (OpenGym, MuJoCo, Atari etc.) \cite{D4RL}. However, the offline learning community has yet to settle to the use of a common pipeline for benchmarks. The results achieved by MOReL using D4RL are reported in Table \ref{table:morel}, for comparison with different baselines examine the reference \cite{MOReL}. Independently of the quality of the batch we notice an improvement in the performance, expressed as the average cumulative reward over a sequence of trajectories, of the optimal policy for the pessimistic MDP when evaluated in the true environment.
 
 The results achieved with MOPO are reported in Table \ref{table:mopo}. MOPO performs better than all previous baselines on randomly generated batches and on data sets which consist of the full replay buffer of a Soft-Actor Critic (SAC) trained partially up to an environmental specific performance threshold. Surprisingly, on batches generated with the sub-optimal trained SAC the best baselines are BRAC with value function penalty and BEAR. The main difference between the last two types of data sets is that while the latter is generated with a fixed policy, the previous one is a collection of transitions gathered with a mixture of differently performing policies. When the sub-optimal policies are not so bad, it seems reasonable  to just slightly modify them to obtain better results, hence BEAR and BRAC looks like viable methods. However, when the overall batch policy is not so good, constraining the reward with respect to the model error (and the transitions close to the ones present in the batch up to a roll-out horizon) can be more fulfilling.

\begin{table}[h]
\resizebox{.95\columnwidth}{!}{
\begin{tabular}{|l|l|l|}
\hline
\textbf{Environment} & \textbf{Pure-Random} & \textbf{Pure-Partial} \\ \hline
Hopper-v2 & $2354 \pm 443$ $(20)$ & $3642 \pm 54$ $(1376)$ \\ \hline
HalfCheetah-v2 & $2698 \pm 230$ $(-638)$ & $6028 \pm 192$ $(4198)$ \\ \hline
Walker2d-v2 & $1290 \pm 325$ $(-7)$ & $3709 \pm 159$ $(1463)$ \\ \hline
Ant-v2 & $1001 \pm 3$ $(-263)$ & $3663 \pm 247$ $(1154)$ \\ \hline
\end{tabular}}
\caption{Average cumulative return of the policy obtained with MOReL as reported in \cite{MOReL}. A Pure-Partial policy is a partially trained suboptimal policy. The number between parentheses is the average cumulative reward with the batch collecting policy. All results are averaged over 5 random seeds.}
\label{table:morel}
\end{table}

\begin{table*}[h!]
\centering
\begin{tabular}{|l|l|l|l|l|l|l|l|l|}
\hline
\textbf{Data set type} & \textbf{Environment} & \textbf{Batch Mean} & \textbf{Batch Max} & \textbf{SAC} & \textbf{BEAR} & \textbf{BRAC-vp} & \textbf{MBPO} & \textbf{MOPO} \\ \hline
random & halfcheetah & -303.2 & -0.1 & 3502.0 & 2885.6 & 3207.3 & 3533.0 & \textbf{3679.8} \\ \hline
random & hopper & 299.26 & 365.9 & 347.7 & 289.5 & 370.5 & 126.6 & \textbf{412.8} \\ \hline
random & walker2d & 0.9 & 57.3 & 192.0 & 307.6 & 23.9 & 395.9 & \textbf{596.3} \\ \hline
medium & halfcheetah & 3953.0 & 4410.7 & -808.6 & 4508.7 & \textbf{5365.3} & 3230.0 & 4706.9 \\ \hline
medium & hopper & 1021.7 & 3254.3 & 5.7 & \textbf{1527.9} & 1030.0 & 137.8 & 840.9 \\ \hline
medium & walker2d & 498.4 & 3752.7 & 44.2 & 1526.7 & \textbf{3734.3} & 582.5 & 645.5 \\ \hline
mixed & halfcheetah & 2300.6 & 4834.2 & -581.3 & 4211.3 & 5413.8 & 5598.4 & \textbf{6418.3} \\ \hline
mixed & hopper & 470.5 & 1377.9 & 93.3 & 802.7 & 5.3 & 1599.2 & \textbf{2988.7} \\ \hline
mixed & walker2d & 358.4 & 1956.5 & 87.8 & 495.3 & 44.5 & 1021.8 & \textbf{1540.7} \\ \hline
med-expert & halfcheetah & 8074.9 & 12940.2 & -55.7 & 6132.5 & 5342.4 & 926.6 & \textbf{6913.5} \\ \hline
med-expert & hopper & 1850.5 & 3760.5 & 32.9 & 109.8 & 5.1 & \textbf{1803.6} & 1663.5 \\ \hline
med-expert & walker2d & 1062.3 & 5408.6 & -5.1 & 1193.6 & \textbf{3058.0} & 351.7 & 2527.1 \\ \hline
\end{tabular}
\caption{Results for D4RL datasets as reported in \cite{MOPO}. Each number is the average undiscounted return of the policy at the last
iteration of training, averaged over 3 random seeds. Data set types depend on the policy used to collect the batch: random (random policy), medium (suboptimally trained agent with a Soft Actor-Critic), mixed (adoperate as batch the replay buffer use to train a Soft-Actor Critic until an envirnomental specific threshold is reached), medium-expert (mix of an optimal policy and a random or a partially trained one). SAC column stands for a Soft Actor Critic (model-free) agent. BRAC-vp is the version of BRAC with the value penalty. MBPO is the vanilla model-based algorithm described in \cite{MBPO}. Check the reference \cite{MOPO} for details about the hyperparameters.}
\label{table:mopo}
\end{table*}

As mentioned also by the authors of BRAC, their algorithm when applied to small data sets becomes more susceptible to the choice of the hyperparameters. This is probably because on small data set the distributional shift / model error can become significant. It is crystal clear that the field needs better theoretical foundations and better algorithms in order to learn more safe and performing policies from small batches collected with strategies of any quality, even uniform random ones.

\section{Conclusions}
\label{sec:conclusion}

\noindent In this paper we have examined the state-of-the-art RL and planning algorithms motivated by the necessity to exploit their application to improve offline learning using a single batch of collected experiences. This is challenging problem of crucial importance for the development of intelligent agents. In particular, when the interaction of such agents with the environment is expensive, risky or unpractical. Our goal was that of providing to the reader a self-contained summary of the general ideas that flow behind the main topic. For simplicity, we focused on MDPs but once the listed difficulties will be addressed we aim to extend the discussion to Partially Observable MDPs which are a more appropriate object to describe the interaction of agents in partial observable environments. %collaboration between robotic and biologic agents.
We started with a recap of MDPs and resolution schemes. % Then we sketched an introduction on the single bacth learning and planning problem} %to state aggregation and state representation learning. 
Then we presented the single batch learning and planning problem.
Our main contribution is an outline of model-free and model-based batch RL algorithms while providing comments on size scalability, efficiency and on the usefulness of theoretical bounds. In particular, we proposed an improvement of the definition of performance of the value function following a specific policy that led us to believing that a sub-optimal policy for a reward uncertainty penalized MDP can be better than the optimal one when applied in the true environment. Secondly, we analyzed the penalization introduced in all sorts of offline-learning algorithms. We showed that if the coefficients multiplying the distributional shift estimator are too big then the theoretical threshold which bounds the performance of the policy applied in the real world is always respected, and therefore of little practical utility. We also advised the future implementation of GANs for a better estimate of distributional shifts and model errors. Indeed, estimators that optimizes a min-max loss function give hint that this might be a viable solution.

\bibliography{main}
\bibliographystyle{aaai}
\end{document}